\documentclass[runningheads]{llncs}
\usepackage{graphicx}
\usepackage{amsmath,amssymb} 
\usepackage{color}                                     

\usepackage{graphics} 
\usepackage{epsfig} 
\usepackage{times} 
\usepackage{afterpage}
\usepackage[final]{pdfpages}
\usepackage{times}
\usepackage{mathtools,enumerate}
\usepackage{multirow}
\usepackage{url}    
\usepackage{cite}
\usepackage{textcomp}
\usepackage{gensymb}
\usepackage{float}
\usepackage{hyperref}
\usepackage{url}
\usepackage{array}
\usepackage{mathtools}
\usepackage{subfiles}
\usepackage{booktabs}
\graphicspath{{figures/}{../figures/}}

\newlength{\depthofsumsign}
\newcommand{\etal}{\mbox{\emph{et al.}}}
\newcommand{\ie}{\mbox{\emph{i.e.}}}
\newcommand{\eg}{\mbox{\emph{e.g.}}}
\newcommand{\tildeg}{\mathit{\tilde{G}}}
\DeclareMathOperator{\diag}{diag}

\newcommand{\nsum}[1][1.5]{
    \mathop{%
        \raisebox
            {-#1\depthofsumsign+1\depthofsumsign}
            {\scalebox
                {#1}
                {$\displaystyle\sum$}%
            }
    }
}

\newcommand{\argmin}[1]{\underset{#1}{\operatorname{arg}\,\operatorname{min}}\;}

\begin{document}

\title{Dealing with Ambiguity in Robotic Grasping \\ via Multiple Predictions} 

\titlerunning{Dealing with Ambiguity in Robotic Grasping via Multiple Predictions} 


\author{
Ghazal Ghazaei\inst{1,2} \and
Iro Laina\inst{2} \and
Christian Rupprecht\inst{2} \and \\
Federico Tombari\inst{2} \and
Nassir Navab\inst{2} \and
Kianoush Nazarpour\inst{1,3}}

\authorrunning{G. Ghazaei et al.}


\institute{School of Engineering, Newcastle University, Newcastle, UK \and 
            Technische Universit\"at M\"unchen, Munich, Germany \and
            Institute of Neuroscience, Newcastle University, Newcastle, UK}
\maketitle
%
\begin{abstract}
Humans excel in grasping and manipulating objects because of their life-long experience and knowledge about the 3D shape and weight distribution of objects. However, the lack of such intuition in robots makes robotic grasping an exceptionally challenging task. There are often several equally viable options of grasping an object. However, this ambiguity is not modeled in conventional systems that estimate a single, optimal grasp position. We propose to tackle this problem by simultaneously estimating multiple grasp poses from a single RGB image of the target object. Further, we reformulate the problem of robotic grasping by replacing conventional grasp rectangles with grasp belief maps, which hold more precise location information than a rectangle and account for the uncertainty inherent to the task. We augment a fully convolutional neural network with a multiple hypothesis prediction model that predicts a set of grasp hypotheses in under 60~ms, which is critical for real-time robotic applications. The grasp detection accuracy reaches over \textbf{$90\%$} for unseen objects, outperforming the current state of the art on this task. 
\end{abstract}

\section{Introduction}
Grasping is a necessary skill for an autonomous agent to interact with the environment.
The ability to grasp and manipulate objects is imperative for many applications in the field of personal robotics and advanced industrial manufacturing. However, even under simplified working conditions, robots cannot yet match human performance in grasping. While humans can reliably grasp and manipulate a variety of objects with complex shapes, in robotics this is still an unsolved problem. This is especially true when trying to grasp objects in different positions, orientations or objects that have not been encountered before. Robotic grasping is a highly challenging task and consists of several components that need to take place in real time: perception, planning and control. 
 
In the field of robotic perception, a commonly studied problem is the detection of viable grasping locations. Visual recognition from sensors ---such as RGB-D cameras--- is required to perceive the environment and transfer candidate grasp points from the image domain to coordinates in the real world. The localization of reliable and effective grasping points on the object surface is a necessary first step for successful manipulation through an end effector, such as a robotic hand or a gripper. The detected target position is then used such that an optimal trajectory can be planned and executed.
This visual recognition task has gained great attention in recent years \cite{saxena2008robotic,jiang2011efficient,lenz2014,redmon2015real,wang2016robot,guo2017hybrid,kumra2016robotic,asif2017rgb,levine2016learning,icinco17,kehoe2015survey,viereck2017learning,varley2016shape,miller2004graspit} and led to the emergence of benchmark datasets, such as the Cornell grasp detection dataset \cite{lenz2014}, to evaluate the performance of approaches designed for this specific task. 

\begin{figure}[t]
    \centering
    \includegraphics[width=0.8\textwidth]{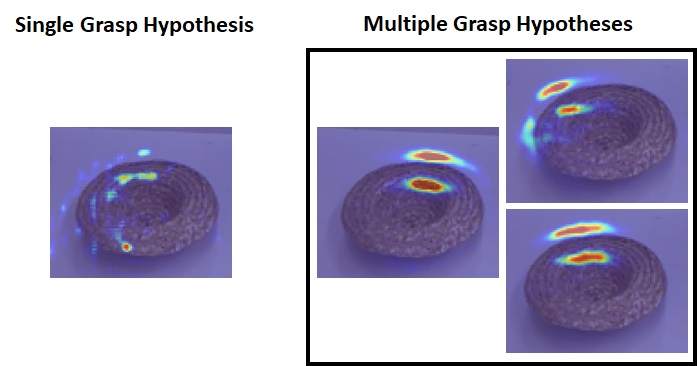}
    \caption{We propose a model for regressing multiple grasp hypotheses as 2D belief maps, which tackles the ambiguity of grasp detection more effectively than a single grasp detection, in particular for completely unseen shapes, as the one depicted here.}
    \label{fig:teaser}
\end{figure}

Early approaches rely on explicitly estimating object geometry to localize grasping points \cite{miller2004graspit,icinco17}. This tends to slow down the overall run-time and fails in presence of complicated or unseen object shapes. Following the success of deep learning in a wide spectrum of computer vision applications, several recent approaches \cite{lenz2014,redmon2015real,wang2016robot,guo2017hybrid,kumra2016robotic,viereck2017learning,mahler2017dex} employed Convolutional Neural Networks (CNNs) \cite{Lecun1998,Krizhevsky2012} to successfully detect grasping points from visual data, typically parametrized by 5-dimensional (5D) grasping representations \cite{jiang2011efficient,lenz2014}. 
It is worth noting that most of these methods rely on depth data, often paired with color information. 
All these approaches have contributed significantly to improving robotic grasp detection, however they have not exhaustively studied generalization to novel, complex shapes.
In particular, although some prior work explicitly aims at grasp estimation for unseen objects from RGB-D/depth data, 
this aspect is still regarded as an open issue \cite{icinco17}.  
In this work we propose a novel grasp detection approach from RGB data only. Our method incorporates two measures to explicitly model ambiguity related to the task of robotic grasping. First, we redefine the task of grasp detection as dense belief estimation problem. 
Thus, instead of the conventional grasp representation based on bounding boxes~\cite{jiang2011efficient} we model the grasp space with 2D belief maps to be predicted from an input image.
This allows the model to predict a grasp distribution with spatial uncertainty that accounts for small-scale ambiguities and exploits the full potential of CNNs in learning spatial representations.

The reformulation of this problem further highlights the inherent ambiguity in grasp detection. Most objects can be gripped in different ways and, although some may be preferable, there is not necessarily a ``best'' grip. 
This is also reflected in that current benchmarks provide multiple grasp rectangles as ground truth for each object. 
However, aiming for a single output in an ambiguous problem can harm performance as the network typically learns the conditional average of all possible outcomes. 
To better model larger scale ambiguities, we employ a \emph{multi-grasp prediction} framework and estimate multiple meaningful grasping positions for each input image. 
This approach allows to better model the output distribution and results in more precise and robust predictions especially in the case of unseen objects. 
The outcome of our method in comparison to a conventional single-prediction model is depicted in Figure~\ref{fig:teaser}.
Finally, for the selection of a single grasping position, we propose an additional ranking stage based on Gaussian Mixture Models (GMMs)~\cite{mclachlan2004finite}. This is particularly useful for practical applications of our approach and for fair comparisons with the state of the art. We demonstrate the effectiveness of our approach by evaluating on a common benchmark \cite{lenz2014}.
\section{Related Work}

\subsubsection{Robotic Grasp Detection }
Before the immense success of deep learning in computer vision applications, grasp estimation solutions were mostly based on analytic methods \cite{bicchi2000robotic}. Some of these approaches, such as Graspit! \cite{miller2004graspit}, are dependent on the presence of a full 3D model to fit a grasp to it, not feasible for real-time applications. With the improvement of depth sensors, there are also recent methods that leverage geometrical information to find a stable grasp point using single-view point clouds \cite{icinco17}.

In addition, the combination of both learning techniques and 3D shape information has led to interesting results. Varley \etal~\cite{varley2016shape}, use a deep learning based approach to estimate a 3D model of the target object from a single-view point cloud and suggest a grasp using 3D planning methods such as Graspit!. Mahler \etal~\cite{mahler2017dex} develop a quality measure to predict successful grasp probabilities from depth data using a CNN. Asif \etal~\cite{asif2017rgb} extract distinctive features from RGB-D point cloud data using hierarchical cascade forests for recognition and grasp detection.

The most recent robotic grasp estimation research is focused solely on deep learning techniques. Lenz \etal~\cite{lenz2014} pioneered the transfer of such techniques to robotic grasping using a two-step cascade system operating on RGB-D input images. A shallow network first predicts high-ranked candidate grasp rectangles, followed by a deeper network that chooses the optimal grasp points. Wang \etal~\cite{wang2016robot} followed a similar approach using a multi-modal CNN. 
Another method~\cite{kumra2016robotic} uses RGB-D data to first extract features from a scene using a ResNet-50 architecture~\cite{he2016deep} and then a successive shallower convolutional network applied to the merged features to estimate the optimal point of grasping. 

Recent work in robotic grasp detection has also built on object detection methods~\cite{ren2015faster, redmon2016you} to directly predict candidate grasp bounding boxes.
Redmon \etal~\cite{redmon2015real} employ YOLO \cite{redmon2016you} for multiple grasp detection from RGB-D images. This model produces an output grid for candidate predictions including the confidence of grasp being correct in each grid cell. This MutiGrasp approach improved the state-of-the-art accuracy of grasp detection up to~$\sim10\%$. However, the results are only reported for the best ranked rectangle and the performance of other suggested grasps is not known.
Guo \etal~\cite{guo2017hybrid} instead propose a hybrid deep network combining both visual and tactile sensing. The multi-modal data is fed into a visual object detection network~\cite{ren2015faster} and a tactile network during training and the features of both networks are concatenated as an intermediate layer to be employed in deep visual network during test.  

\subsubsection{Landmark Localization } In our method we define the grasping problem differently. Instead of approaching the task as object detection, \ie~detecting grasping rectangles as for example in \cite{guo2017hybrid,redmon2015real}, we express the rectangles as 2D belief maps around the grasping positions. This formulation is inspired by the latest methods in landmark localization, for example in human pose estimation~\cite{belagiannis2017recurrent, bulat2016human, cao2017realtime, papandreou2017towards, wei2016convolutional}, facial keypoint detection \cite{Bulat_2018_CVPR,Merget_2018_CVPR} and articulated instrument localization \cite{du2018articulated,laina2017concurrent}. The use of heat maps to represent 2D joint locations has significantly advanced the state of the art in the localization problem. These models are trained so that the output matches the ground truth heat maps, for example through $\mathcal{L}_2$ regression, and the precise landmark locations can be then computed as the maxima of the estimated heat maps. 

\subsubsection{Multiple Hypothesis Learning} To better model the grasp distribution of varying objects as well as grasp uncertainty, we augment the belief maps along the lines of multiple hypothesis learning~\cite{rupprecht2017learning, lee2016stochastic}.
These methods model ambiguous prediction problems by producing multiple possible outcomes for the same input. 
However, they do not explore the possibility to select the best hypothesis out of the predicted set. 
The problem of selecting good hypotheses for multi-output methods has been typically addressed by training selection networks \cite{Li_2018_CVPR,guzman2014multi}.
Here, we solve this problem in a task-specific fashion, by scoring the predictions based on their alignment with a parametric Gaussian distribution which was used in training. 
\section{Methods}
In the following, we describe our approach in detail. First, we redefine the problem of robotic grasp detection as prediction of 2D grasp point belief maps (Section \ref{subsec:beliefmaps}). Specifically, we learn a mapping from a monocular RGB image to grasping confidence maps via CNN regression (Section \ref{subsec:CNN}). We then introduce our multi-grasp framework to tackle the inherent ambiguity of this problem by predicting multiple grasping possibilities simultaneously (Section \ref{subsec:MHP}). Finally, we rank all predicted grasps according to GMM likelihood in order to select the top ranked prediction (Section \ref{subsec:GMM}).

\begin{figure}[t]
      \centering
      \includegraphics[width=0.60\textwidth]{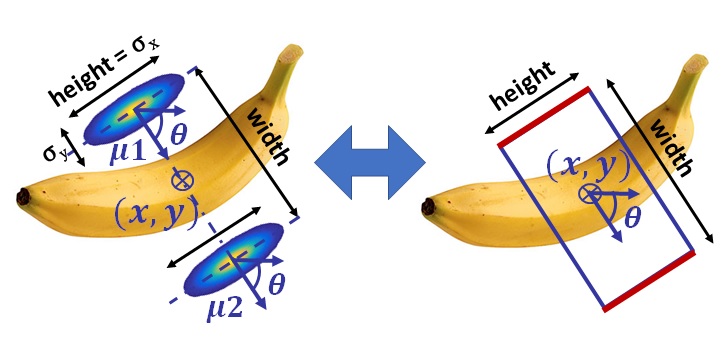}
      \vspace{-1em}
      \caption{An illustration of the adaptation of grasp rectangles to their associated grasp belief maps. The belief maps are constructed using the centers of the gripper plates as means for the normal distributions. The variance $\sigma_x$ is proportional to the gripper height, while $\sigma_y$ is a chosen constant.}
      \label{fig:box_details}
\end{figure}

\begin{figure}[ht]
	\centering
	  \includegraphics[width=0.95\textwidth]{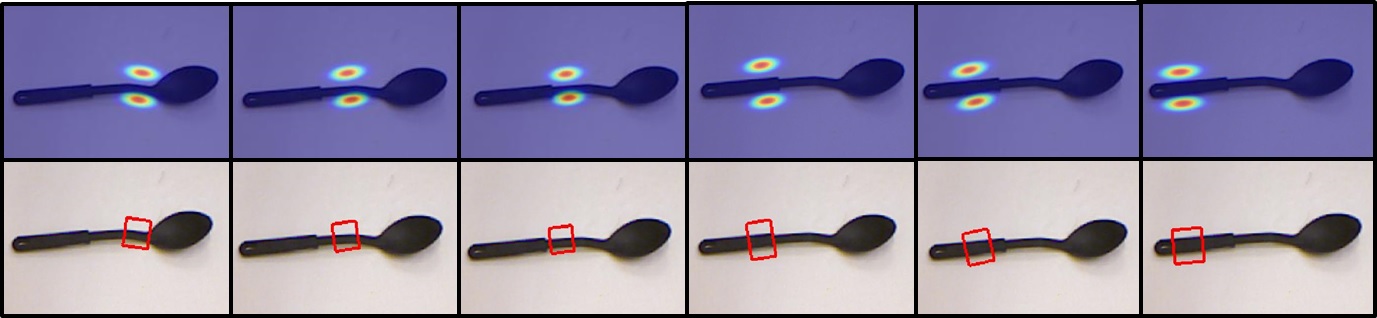}
	  \caption{Samples of rectangle grasps and grasp belief maps shown for the same item.}
    \label{fig:beliefmap}
\end{figure}

\subsection{Grasp Belief Maps} \label{subsec:beliefmaps}
The problem of robotic grasp detection can be formulated as that of predicting the size and pose of a rectangle, which, as suggested by \cite{jiang2011efficient}, includes adequate information for performing the grasp; that is a 5-dimensional grasp configuration denoted by $(x,y,\theta,h,w)$, where $(x,y)$ is the center of the rectangle and $\theta$ is its orientation relative to the horizontal axis. We denote width and height of the bounding box with $w$ and $h$ respectively. These correspond to the length of a grip and the aperture size of the gripper. This representation has been frequently used in prior work~\cite{lenz2014, wang2016robot,redmon2015real,asif2017rgb,kumra2016robotic,guo2017hybrid} as guidance for robotic grippers.

\noindent 
In this work, we propose an alternative approach to model the detection of a robotic grasp using 2D belief maps. For an $N$-finger robotic gripper, these belief maps can be represented as a mixture model of $N$ bivariate normal distributions fitted around the finger locations.

For a parallel gripper, the previously used grasping rectangle representation can be encoded in belief maps as follows. The means $\vec{\mu}^{(n)} = (\mu_x^{(n)}, \mu_y^{(n)})^T$, with $n\in\{1, 2\}$, around which the Gaussian distributions are centered correspond to the 2D centers (in Cartesian coordinates) of the gripper plates. The distance of the means $\|\vec{\mu}^{(1)} - \vec{\mu}^{(2)}\|_2 = w$ represents the width of the grasp. The Gaussian distributions are elliptical with $\Sigma~=~\diag(\sigma_x^{(n)}, \sigma_y^{(n)})^2$. The primary axis of the ellipse represents the grasp height $h$. The orientation of the Gaussian kernels is adjusted by the rotation matrix $R(\theta)$ to make up for the correct grasping pose with respect to the object. The mixture model can be then defined as
\begin{equation}
\label{eq:2DGaussians}
\mathit{G}(\mathbf{p}) = \nsum_{n=1}^{N}
  \frac{
    \exp{
      \Big(
        -\frac{1}{2}
        \big(\mathbf{p}-\vec{\mu}^{(n)}\big)^T 
        R(\theta) \, \Sigma^{-1} \, R(\theta)^T
        \big(\mathbf{p}-\vec{\mu}^{(n)}\big)
      \Big)}}{2\pi N \, \sigma_x^{(n)} \sigma_y^{(n)} }, 
\end{equation}

\noindent 
where $\mathbf{p}$ denotes a pixel's location inside the belief map.
An illustration of our adapted grasp belief maps is shown in Figure~\ref{fig:box_details}. 

Grasp belief maps enclose the same information as the grasp rectangles, while expressing an encoding of the inherent spatial uncertainty around a grasp location. The proposed representation encourages the encoding of image structures, so that a rich image-dependent spatial model of grasp choices can be learned. Moreover, the amplitude as well as variance of the predicted belief maps can act as a measure of confidence for the exact location and orientation of the grasp. In Figure~\ref{fig:beliefmap}, we show all possible grasp configurations for an item using both the traditional bounding box representation and our adapted continuous approach based on belief maps. 

A model equipped with grasp belief maps can express its uncertainty spatially in the map, while direct regression of rectangles makes it harder to model spatial uncertainty.
Further, such mixture models can be seamlessly extended to model grasp representations of other, more complex types of grippers, such as hand prostheses. 

In practice, we create heat maps by constructing Gaussian kernels according to Equation~\ref{eq:2DGaussians}, parametrized by the centers and dimensions of the gripper fingers. The centers of the gripper plates specify the means of the Gaussian kernels, $\sigma_x$ is proportional to the gripper height and $\sigma_y$ is a chosen constant value.

\subsection{CNN Regression} \label{subsec:CNN}
\begin{figure}[t]
      \centering
      \includegraphics[width=0.9\textwidth]{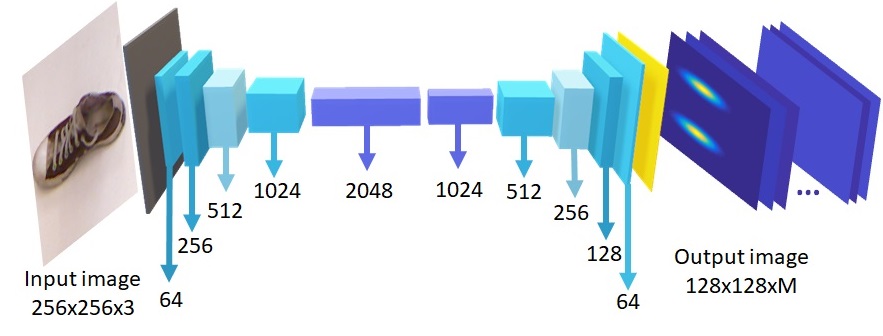}
      \caption{The architecture of the fully convolutional residual network used in this paper. M refers to the number of grasp map predictions.}
      \label{fig:architecture}
\end{figure}

For the regression of confidence maps, a common design choice among deep learning methods have been fully convolutional networks (FCNs) \cite{long2015fully}. For our purpose, we use the fully convolutional residual architecture proposed in \cite{laina2016deeper}, which has shown competitive performance for dense prediction tasks, in particular depth estimation, in real time. The encoder is based on ResNet-50 \cite{he2016deep}, which embeds the input into a low dimensional latent representation. The decoder is built from custom residual up-convolutional blocks, which increase the spatial resolution up to half of the input resolution. The architecture is shown in Figure \ref{fig:architecture}.

Given our problem definition, the network is trained to perform a mapping from a monocular RGB input to a single-channel heatmap comprised of the Gaussian mixture which represents the grasp belief. Since there are typically more than one viable grasp per object, choosing a single ground truth grasp becomes an ambiguous problem. When training in the single-grasp setup, we choose the most stable available grasp as ground truth, that is the one with the maximum grasping area. To this end, the objective function to be minimized is the Euclidean norm between the predicted belief map $\tilde{\mathit{G}}$ and the chosen ground truth map:
\begin{equation}\label{eq1}
\mathcal{L(\tilde{\mathit{G}}, \mathit{G})} = \|\mathit{\tilde{G}} - \mathit{G}\|_2^2
\end{equation}

\subsection{Multiple Grasp Predictions} \label{subsec:MHP}

Training the model with a single viable grasp is not optimal and could harm generalization, because the model gets penalized for predicting grasps which are potentially valid, but do not exactly match the ground truth. In other words, the samples that the model learns from do not cover the entire grasp distribution. Thus, in the case of known objects, the model would overfit to the single grasp possibility it has seen, while in the case of previously unseen objects the uncertainty which arises would prevent the model from producing a sharp and reliable belief map (Figure~\ref{fig:teaser}).

To overcome this limitation we propose a multi-grasp estimation setup, 
Instead of forcing the model to produce exactly one grasp, we allow the same model to produce multiple simultaneous outputs $\tildeg = \{\tildeg^{(m)}\}_m$, $m \in \{1,2,...,M\}$. In practice, we replicate the last layer $M$ times. Our goal is to then train the model such that it approximates the entire distribution of viable grasps. This problem can be formulated as an oracle meta-loss $\mathcal{M}$ that acts on top of the problem-specific objective function $\mathcal{L}$. By denoting the cost value of each grasp output as 
\begin{equation}
 \mathcal{L}_m = \mathcal{L}(\tildeg^{(m)}, \mathit{G})
\end{equation}

\noindent we can then define the meta-loss through the following minimum formulation:
\begin{align}\label{eq:mhp_loss}
    \mathcal{M} (\tildeg,\mathit{G}) &= 
    (1 -{\epsilon})\min_{m = 1, \ldots, M}\mathcal{L}_m + \frac{\epsilon}{M-1} \sum_{m' \ne \argmin{m} \mathcal{L}_{m}} \mathcal{L}_{m'} 
\end{align}
\noindent
The proposed algorithm works as follows. At each training step, a grasp belief map is chosen randomly as the ground truth label among all available ground truth possibilities for the given input sample.  In this way, the entire grasp distribution for each sample will be seen during training. 
Since the model cannot know which ground truth belief map will be chosen for a specific image, it will learn to disentangle the possibilities into the $M$ grasping hypotheses.
This is achieved by the loss $\mathcal{M}$ in Equation \ref{eq:mhp_loss}. This objective is based on the hindsight loss, which only considers the output $\tildeg^{(m)}$ which is closest to the given ground truth $\mathit{G}$. Here we formulate it in a more intuitive way by using a soft approximation in which the oracle selects the best grasp with weight $1-\epsilon$ and $\dfrac{\epsilon}{M-1}$ for all the other predictions, where $\epsilon = 0.05$. This enables the output branches to be trained equally well, especially if they were not initially selected.

\subsection{Grasp Option Ranking} \label{subsec:GMM}
Our previously described model predicts $M$ grasp hypotheses.
For this system to be used in practice, we need a method to assess the hypotheses quality and find which one should be selected. Therefore, it is desirable to find a way to rank all candidate grasps and pick one with a high probability of successful grasping. 
As we train the model to produce two multivariate normal distributions, one way to rank the predicted belief maps is by fitting a two-component Gaussian mixture model (GMM) to each output map using finite mixture model estimation~\cite{mclachlan2004finite}. 

The main parameters of a Gaussian mixture model are the mixture component weights \(\phi_k\) and the component means \(\mu_k\) and variances/covariances \(\sigma_k\) 
with $K$ being the number of components. 
The mathematical description of a GMM distribution over all the components is

\begin{equation}\label{eq:gmm}
p(\vec{x}) = \sum_{k=1}^K\phi_k \mathcal{N}(\vec{x} \;|\; \vec{\mu}_k, \sigma_k),  \quad
\sum_{k=1}^K\phi_k = 1\\
\end{equation}

\noindent where $\mathcal{N}(\vec{x} \;|\; \vec{\mu}, \sigma)$ represents a normal distribution with mean \(\mu\) and variance \(\sigma\).
Mixture models can be estimated via the expectation maximization (EM) algorithm~\cite{dempster1977maximum
}, as finding the maximum likelihood analytically is intractable. EM iteratively finds a numerical solution to the maximum likelihood estimation of the GMM. The EM algorithm follows two main steps: (E) computes an expectation of component assignments for each given data point given the current parameters and (M) computes a maximum likelihood estimation and subsequently updates the model parameters. The model iterates over E and M steps until the error is less than a desired threshold.

We fit the same parametric model that was used to create the ground truth belief maps (Equation \ref{eq:2DGaussians}) and use the likelihood of the fit for each of the $M$ predictions for ranking and choose the best fitted prediction as the system's final output.

\section{Experiments and Results}
In this section, we evaluate our method experimentally on a public benchmark dataset and compare to the state of the art. Further, we investigate the influence of the number of grasp hypotheses $M$ on the performance of the method. 

\subsection{Dataset}
\subsubsection{Cornell dataset } 
\begin{figure}[t]
       \centering
      \includegraphics[width=0.6\textwidth]{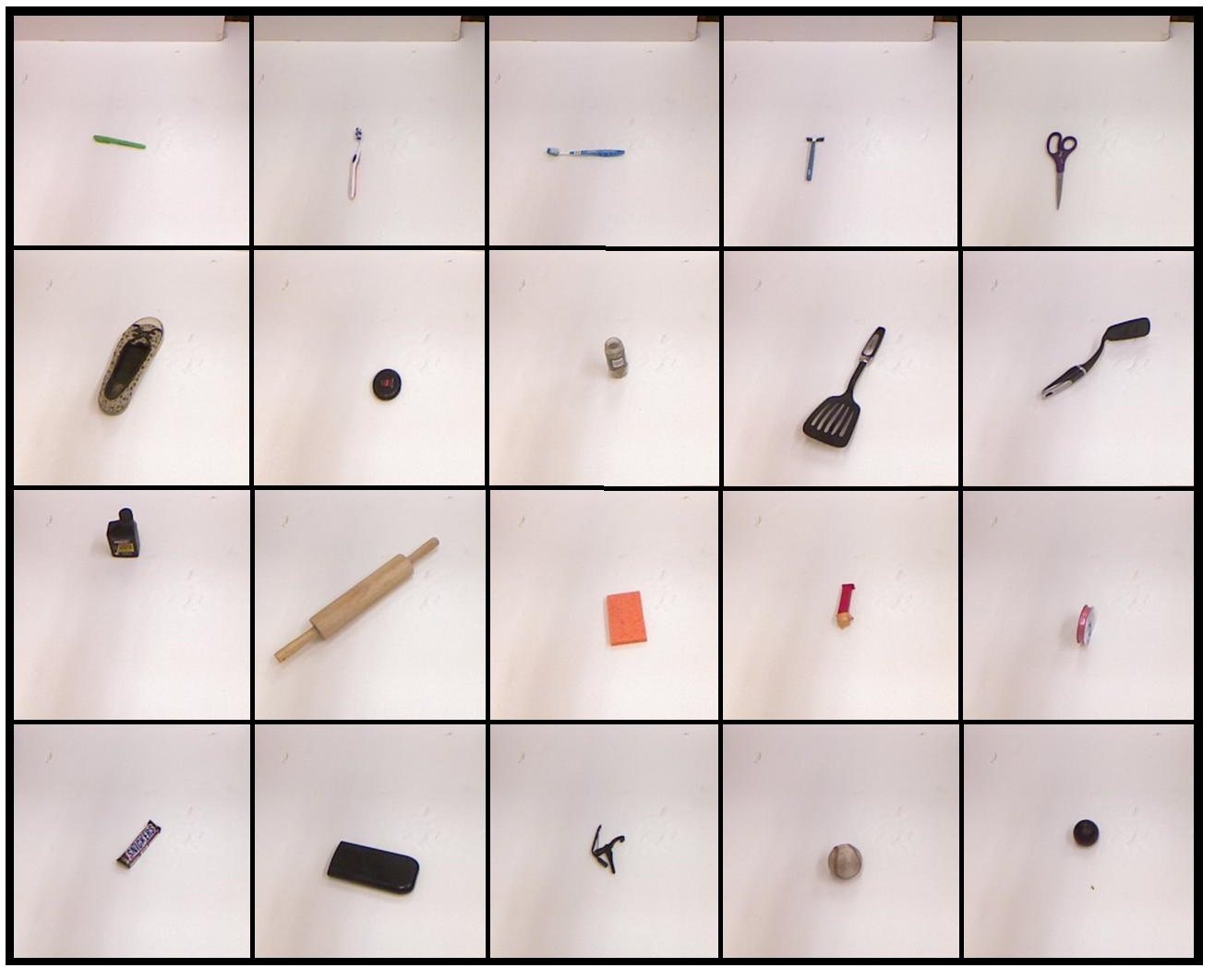}
      \caption{A representation of a subset of the objects of the Cornell grasp detection dataset~\cite{lenz2014}.}
      \label{fig:dataset}
\end{figure}

We evaluate our approach on the Cornell grasp detection dataset~\cite{lenz2014}, which consists of 885 RGB-D images from 240 graspable objects with a resolution of $640\times480$ pixels. The annotated ground truth includes several grasping possibilities per object represented by rectangles.
The dataset is mainly suited for 2D grippers with parallel plates, but as the grasp size and location are included in the representation, it has the potential to be used also for other types of grippers as it is used in~\cite{guo2017hybrid} for a 3-finger gripper. There are 2 to 25 grasp options per object of different scales, orientations and locations, however, these annotated labels are not exhaustive and do not contain every possible grasp. Figure~\ref{fig:dataset} shows some cropped samples of the dataset as used in this work. Here we only use the RGB images and disregard the depth maps. 

\subsubsection{Data splits } 
We follow a cross-validation setup as in previous work~\cite{lenz2014,wang2016robot, redmon2015real, asif2017rgb, kumra2016robotic, guo2017hybrid}, using image-wise and object-wise data splits. The former split involves training with all objects, while some views remain unseen to evaluate the intra-object generalization capability of the methods. However, even an over-fitted model could perform well on this split. The object-wise split involves training on all available views of the same object and testing on new objects and thus is suitable for evaluating inter-object performance. However, the unseen objects are rather similar to ones used in training. 

It is worth noting that none of the previous methods studied the potential of generalizing to truly novel \emph{shapes}, as the dataset includes a variety of similar objects. For example, there are several objects with different colors but of the same shape. Therefore, the object-wise split may not be a good measure for generalization to novel shapes. To investigate our framework's performance on unseen shapes, we have created an additional \emph{shape-wise} split, to encourage larger variation in objects between the train and test sets. We pick the train and test folds such that all the objects of similar shapes, \eg~various kinds of hats, are included in one of the test/train folds only and therefore novel when testing. Both image-wise and object-wise splits are validated in five folds. We perform two-fold cross validation for the shape-wise split, where we use the first 20\% of objects for testing and the remainder for training. The second fold uses the same split but with reversed order of objects.

\subsection{Implementation details} \label{sec:experimental_setup}
In all our experiments we pre-process the images and annotations as detailed in the following.
As the images contain a large margin of background around the objects, we crop them and their corresponding grasp maps to $350\times350$ pixels and then bilinearly down-sample the image to $256\times256$ and the grasp map to $128\times128$.
Prior to cropping we apply data augmentation techniques. We sample a random rotation in $[{-60}^{\circ} , 60^{\circ}]$, center crops with a translation offset of $[-20,  20]$ pixels and scaling between $0.9$ and $1.1$. Each image is augmented six times. Thus, the final dataset contains $5310$ images after augmentations. All the images and labels are normalized to a range of $[0 , 255]$.

To train the single grasp prediction model, we choose the largest ground truth grasp rectangle as label since area is a good indicator for probability and stability of the grasp. This selection may be trivial, but training a single grasp model is not feasible without pre-selection of a fixed ground truth among the available grasps. 

On the other hand, our multiple grasp prediction model can deal with a \textit{variable} number of ground truth grasp maps per image. At each training step, we randomly sample one of the available ground truth annotations. We also add hypothesis dropout with rate $0.05$ as regularization \cite{rupprecht2017learning}. We investigate and report the performance of our framework for different numbers $M$ of grasp hypotheses. To rank multiple predicted grasps, we performed EM steps for up to 1000 iterations and calculated the negative log-likelihood for the parameters $\sigma_k$ and $\mu_k$.

Training was performed on an NVIDIA Titan Xp GPU using MatConvNet~\cite{vedaldi15matconvnet}. The learning rate was set to $0.0005$ in all experiments. For regularization we set weight decay to $0.0005$ and add a dropout layer with rate equal to $0.5$. The models were trained using stochastic gradient descent with momentum of $0.9$ for 50 epochs and a batch size of 5 and 20 for training multiple and single prediction models respectively. 

\subsection{Grasp Detection Metric}

We report quantitative performance using the rectangle metric suggested by~\cite{jiang2011efficient} for a fair comparison. A grasp is counted as a valid one only when it fulfills two conditions:
\begin{itemize}
\item The intersection over union (IoU) score between the ground truth bounding box ($B$) and the predicted bounding box ($B^*$) is above $25\%$, where
\begin{equation}\label{eq:iou}
\text{IoU} = \dfrac{B \cap B^*}{B \cup B^*}
\end{equation}

\item The grasp orientation of the predicted grasp rectangle is within $30^\circ$ of that of the ground truth rectangle.
\end{itemize}

This metric requires a grasp rectangle representation, while our network predicts grasp belief maps. We therefore calculate the modes $\mu_1$ and $\mu_2$ as the centers of each elliptical Gaussian for every predicted belief map. The Euclidean distance between these modes should be equal to the grasp rectangle's width $w$ (Figure~\ref{fig:box_details}). 
We compute the height $h$ of the grasp rectangle as the major axis of the ellipse (after binarization of the belief map with a threshold of 0.2). 
We determine the gripper's orientation $\theta$ by calculating the angle of the major axis as $\arctan{\frac{d_1}{d_2}}$; where $d_1$ and $d_2$ are the vertical and horizontal distance between the centers of elliptical Gaussian maps respectively. 
We can then convert the belief maps to a grasping rectangle representation. Under high uncertainty, \ie when a grasp map is considerably noisy, we discard the hypothesis as a rectangle cannot be extracted.
We note that a valid grasp meets the aforementioned conditions with respect to \emph{any} of the ground truth rectangles and compute the percentage of valid grasps as the \textit{Grasp Estimation Accuracy}. 

\begin{table}[t]
\caption{Comparison of the proposed method with the state of the art. \emph{multiple} refers to our multiple prediction models, while \emph{multiple~/~reg} are the models trained with diversity regularization.}
\label{tab:sota}
\begin{center}
\begin{tabular}{l l@{\hskip 1.5em} l@{\hskip 1em} c@{\hskip 1em} c@{\hskip 1em} c}
\toprule
 & & & \multicolumn{3}{c}{Grasp Estimation Accuracy ($\%$)}   \\ 
\multicolumn{2}{l}{Method} & Input & Image-wise & Object-wise & Shape-wise \\ 
\midrule
\multicolumn{2}{l}{Lenz et al. \cite{lenz2014}} & RGB-D & $73.9$ & $75.6$& - \\
\multicolumn{2}{l}{Wang et al.\cite{wang2016robot}} & RGB-D &  $85.3$ & -  & - \\
\multicolumn{2}{l}{Redmon et al.\cite{redmon2015real}} & RGB-D & $88.0$ & $87.1$& - \\ 
\multicolumn{2}{l}{Asif et al.\cite{asif2017rgb}} & RGB-D &  $88.2$ & $87.5$  & - \\
\multicolumn{2}{l}{Kumra et al.\cite{kumra2016robotic}} & RGB-D &  $89.2$ & $89.0$  & - \\
\multicolumn{2}{l}{Guo et al.\cite{guo2017hybrid}} & RGB-D, tactile & $\mathbf{93.2}$ & $89.1$ & -  \\ 
\midrule
\multicolumn{2}{l}{Kumra et al.\cite{kumra2016robotic}} & RGB &  $88.8$ & $87.7$ & -  \\
\emph{single} & $M = 1$ & RGB & $83.3$ & $81.0$ & $73.7$  \\ 
\emph{multiple} & $M = 5$ & RGB & $91.1$ & $\mathbf{90.6}$ & $85.3$  \\ 
\emph{multiple~/~reg} & $M = 5$ & RGB & $89.1$ & $89.2$ & $82.5$  \\
\emph{multiple} & $M = 10$ & RGB & $\mathbf{91.5}$ & $90.1$ & $\mathbf{86.2}$  \\
\bottomrule
\end{tabular}
\end{center}
\end{table}

\subsection{Evaluation and Comparisons}
In the following, we compare our multiple grasp prediction method with the single-grasp baseline and state-of-the-art methods. As there are several ground truth annotations per object, we compare the selected prediction to all the ground truth grasp rectangles to find the closest match. Among the predictions there can be some which are not viable, while others are perfect matches. The selected prediction for each image is one with the maximum GMM likelihood. 

\subsubsection{Quantitative results} 
We report the results in Table~\ref{tab:sota}, where $M$ indicates the number of hypotheses and consequently $M=1$ refers to the regression of single belief map and can be seen as a baseline in the following experiments. The proposed model with $M=5$ predicted grasps shows significant improvement in performance over the single-grasp model (the average number of grasps per object in the dataset is also approximately five). This performance boost reveals the potential of modeling ambiguity in robotic grasp detection. To study the effect of the number of grasping options, we also evaluated our approach with $M=10$.

While it only relies on RGB data as input, our multiple grasp approach outperforms all state-of-the-art methods that use additional depth information, except for Guo \etal~\cite{guo2017hybrid} who also leverage tactile data. Moreover, both single and multiple grasp models have a faster grasp prediction run-time than the state of the art at $56$ ms. GMM maximum likelihood estimation for hypothesis selection increases the run-time to $95$ ms. Increasing the number of outputs $M$ does not have a negative effect on speed.

It is worth noting that the comparable performance of the models in the image- and object-wise splits (also in prior work) suggests that task difficulty does not change much between the two scenarios. With the more challenging shape-wise scenario that we have proposed, we can better evaluate performance on novel objects. In this case, the accuracy of the single grasp baseline drops significantly. On the contrary, the multiple grasp model is still able to handle the increased difficulty with a large performance boost over the baseline. 
It can be observed that with an increasing number of grasp hypotheses the performance gap of the multiple-grasp over the single-grasp model is the highest for the shape-wise split, with over $12\%$ increase in accuracy for unseen shapes/objects. 

\begin{figure}[t]
      \centering
      \includegraphics[width=\linewidth]{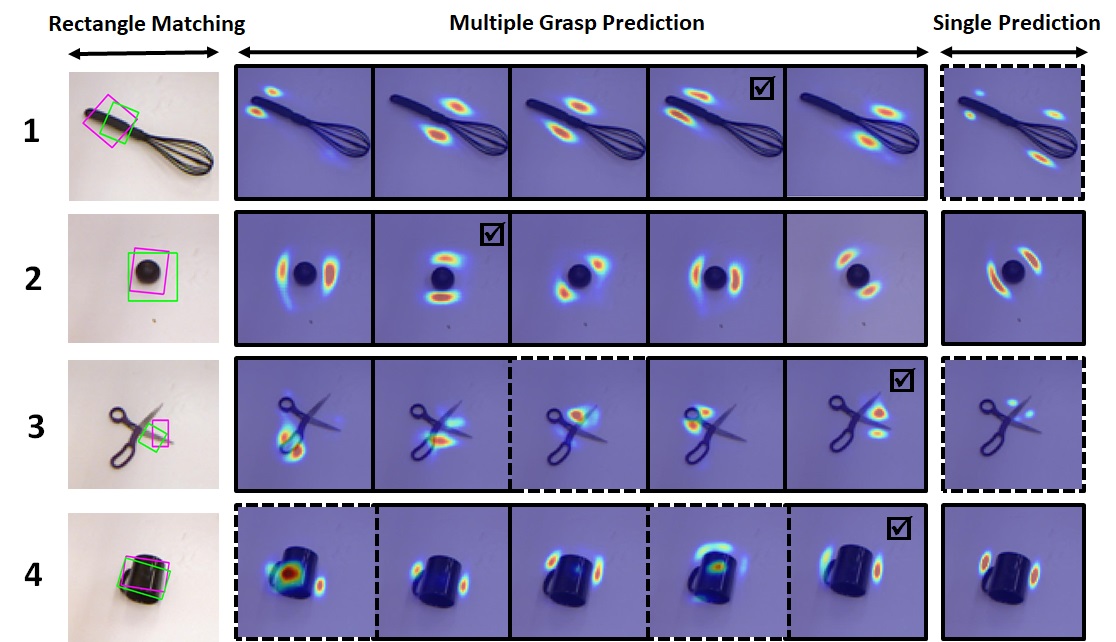}
      \caption{Five and single grasp map predictions of sample objects in the dataset. A solid frame around an image is an indicator of grasp detection success, while a dashed line shows an incorrect detection. The predictions marked with $\checkmark$ are the top-ranked ones according to the GMM likelihood estimation. These predictions are converted back to grasp rectangles (magenta) and compared with the closest ground truth grasp (green).}
      \label{fig:qualitative}
\end{figure}

\begin{figure}[t]
      \centering
      \includegraphics[width=0.95\linewidth]{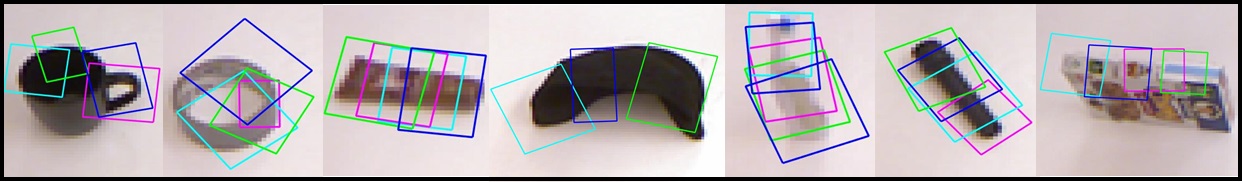}
      \caption{Examples of diversity within predicted grasp maps (converted to rectangles).}
      \label{fig:diversity}
\end{figure}

\subsubsection{Diversity of predictions}
We also examine the diversity of the predicted hypotheses for each image. 
We have performed experiments adding a repelling regularizer~\cite{rochan2018video} (weighted by a factor of $0.1$) to the hindsight loss to further encourage diverse predictions. 
The accuracy of this model with $M=5$ (Table~\ref{tab:sota}) is slightly worse than our multiple prediction model without the regularizer. 
As a measure of hypothesis similarity, we calculate the average cosine distance among all predictions given an input. The average similarity for the object-wise split decreases only marginally from 0.435 (without regularizer) to 0.427 (with regularizer), suggesting that the multiple prediction
framework does not really benefit by explicitly optimizing diversity. Our framework can naturally produce diverse predictions, which we intuitively attribute to the hypothesis dropout regularization used during training. 

\subsubsection{Qualitative results } 
In Figure~\ref{fig:qualitative} we show qualitative examples from our multi-grasp framework (with $M=5$) and a comparison to the single grasp ($M=1$) model's predictions, noting the advantage of multiple grasp predictions both in terms of accuracy and variability. We observe that for objects that have several distinct grasping options, our multiple prediction framework models the output distribution sufficiently. Object 3 (scissors) is undoubtedly a challenging object with many different grasping poses, which are successfully estimated via multiple predictions. 
In Figure~\ref{fig:diversity} we further emphasize the diversity among the grasp hypotheses, by showing multiple extracted rectangles for various objects.

\begin{table}[t]
\caption{Average grasp estimation accuracy of all hypotheses (lower limit) and average grasp success (upper limit).}
\label{tab:ablation}
\begin{center}
\begin{tabular}{lr@{\hskip 1em}r@{\hskip 1em}r}
\toprule
Method & Image-wise & Object-wise & Shape-wise\\ \midrule
lower limit ($M=5$) & $80.0$ & $77.4$ & $75.0$  \\ 
lower limit ($M=10$)& $76.5$ & $73.3$ & $72.1$  \\  \midrule
upper limit ($M=5$)& $98.0$ & $98.5$ & $96.3$  \\ 
upper limit ($M=10$)& $99.2$ & $98.4$ & $99.1$  \\ 
\bottomrule
\end{tabular}
\end{center}
\end{table}

\subsection{Evaluation over Multiple Grasps}
In Table~\ref{tab:ablation} we report the lower and upper detection accuracy limits of the multi-grasp models.
Instead of evaluating only the top-ranked grasp hypothesis, we first evaluate \emph{all} predictions provided by our model. This evaluation gives the lower limit of the model's performance, as it computes the success rate of all hypotheses, including even those with a low probability of being chosen. This result suggests that the estimated belief maps correspond, in most cases, to valid grasps ($75\%$ overall accuracy compared to $85.3\%$ for one chosen grasp in shape-wise split, when $M=5$). This lower bound decreases as $M$ increases, \ie~it is more likely to have a (noisy) prediction that does not match any of ground truth grasp rectangles with higher $M$. However, thresholding the ``good" matches based on GMM likelihood can counteract this drop in performance while leaving multiple grasping choices to the robot. 

Another observation is that the top-ranked prediction is not necessarily the best one in terms of grasping performance. This can be seen in the upper limit evaluation, in which if there exists at least one matching grasp detection among all hypotheses, it counts overall as successful. For $M=10$ the upper limit exceeds $98\%$ accuracy for the object-wise split. This implies that there is in almost all cases at least one valid prediction returned by our model, although GMM fitting might not always result in correct ranking. Still, the top-ranked prediction performance in Table~\ref{tab:sota} is closer to the upper rather than the lower limit.  

\subsection{Generalization}
\begin{figure}[t]
      \centering
      \includegraphics[width=0.9\textwidth]{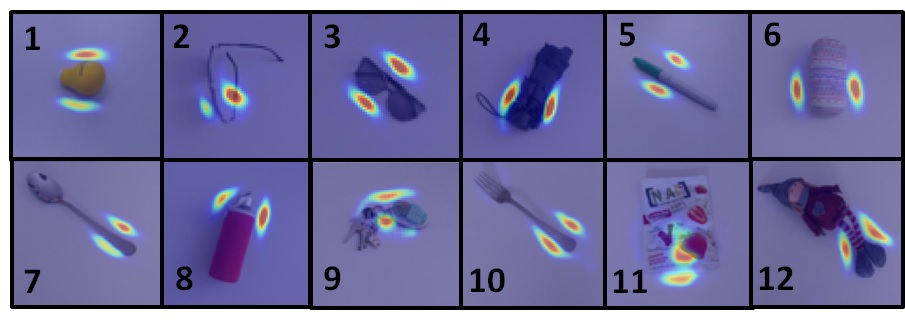}
      \caption{The top ranked grasp map selected by the GMM likelihood estimation module for a $M=5$ model evaluated on common household objects in real-time. Objects 1-5 have similar shapes to the objects in the Cornell grasp dataset. Objects 6-12, however, represent novel shapes and textures compared to the dataset used for training. Despite variations from the training distribution, our method produces reasonable grasp maps for all tested objects.}
      \label{fig:realobjs}
\end{figure}

Finally, to evaluate the performance of the proposed model in a real-world scenario, we test it on several common household objects, such as cutlery, keys and dolls, in an own setup ---and not test images from the same dataset. The differences to the Cornell dataset are not only in the type of objects used, but also in the camera views and illumination conditions. Through this setup we evaluate the generalization ability of the model under different conditions and challenging novel shapes and textures. Figure~\ref{fig:realobjs} illustrates the evaluated objects and the estimated grasp that is chosen as the maximum GMM likelihood. Our model is robust against these variations and results in viable and confident grasping options for all tested objects. 

\section{Conclusion}

We have developed an efficient framework for robotic grasp detection. 
The representation of a grasp is redefined from an oriented rectangle to a 2D Gaussian mixture belief map that can be interpreted as the confidence of a potential grasp position. 
This allows us to handle the ambiguity stemming from the many possible ways to grasp an object.
We employ a fully convolutional network for belief map regression and estimate a variety of viable grasp options per object. 
This approach embraces the ambiguous nature of the grasping task and provides a better approximation of the grasp distribution. 
This property manifests itself in the majority of the predicted grasps being viable solutions and the improvement over the single-grasp baseline becoming larger when tackling scenarios with increased difficulty, such as novel objects, shapes and textures. 
Our ranking approach selects the grasp positions with the highest likelihood, which result in real-time, state-of-the-art performance. 
Considering the fact that our belief map formulation also contains a measure of size, an interesting future direction could be the application of this method to prosthetic hands. 
\subsubsection*{Acknowledgments}
This work is supported by UK Engineering and Physical Sciences Research Council (EP/R004242/1). We also gratefully acknowledge the support of NVIDIA Corporation with the donation of a Titan Xp GPU used for the experiments.

\bibliographystyle{splncs04}
\bibliography{references} 

\end{document}